# Design And Flight Testing Of LQRi Attitude Control For Quadcopter UAV

Astik Srivastava
*Department of Applied Physics*
Delhi Technological University
New Delhi, India
41.astiksrivastava@gmail.com

Richa Sharma
*Department of Applied Physics*
Delhi Technological University
New Delhi, India
richasharma@dtu.ac.in

S. Indu
*Department of Electronics and*
*Communications Engineering*
Delhi Technological University
New Delhi, India
s.indu@dtu.ac.in

*Abstract*—This paper presents the design, implementation, and flight test results of linear quadratic integral regulator (LQRi) based attitude control for a quadcopter UAV. We present the derivation of the mathematical model for the kinematics and dynamics of the UAV, along with the linearized state space representation of the system about hover conditions. LQR and LQRi controllers are then designed to stabilize the UAV in hover conditions and to track desired attitude commands. The controllers are then implemented onboard the Pixhawk flight controller and flight test results are discussed. Finally, the code related to this paper has been published open-source for replication and further research.

## I. Introduction

In recent years, quad-copter UAVs have become immensely popular among hobbyists and researchers alike. They provide a mechanically simple design, heavily used for aerial photography, inspection [1], surveying, surveillance [2], etc. The critical aspect of these systems is their attitude control, which ensures stability and precise maneuverability in diverse operational, and cluttered environments [3]. Over the years, various control strategies have been developed for quadcopter UAVs. PID [4] based control is the most popular approach used for attitude control. PID controllers are intuitive and have a computationally cheap implementation. However, the controllers require tuning of three parameters for each axis, which is cumbersome. Sliding mode control [5], model predictive control [6], [7], dynamic inversion control [8], backstepping control [9], etc are some of the other approaches commonly used for Quadcopter attitude control.

Linear Quadratic Regulator is a control strategy that is commonly used in aerospace applications. LQR is based on optimal control theory, i.e. given the dynamics of a linear system, along with the associated penalties for difference between actual and desired states and control actions, it can be used to create a full-state feedback gain matrix, that optimally regulates the states. Multiple studies have been conducted on the application of LQR for UAV control. [10] proposed an LQR attitude control based on quaternions. In [11], the attitude controller is cascaded into attitude and attitude rate controllers, and an LQR controller is designed to track body rate commands. This enables the quadcopter to track agressive trajectories with greater accuracy. However, this relies on knowledge of a mathematical model for motors used in the UAV, to estimate the thrust being produced by each motor. As the speed of UAV increases, nonlinear effects in the system due to aerodynamic effects increases. To counter this, [12] proposes a state-dependent LQR controller for generating reference body-rate commands. The dynamics are linearized and LQR gain is calculated continously, which is then used to generate reference body rates for a low-level controller. Although this allows the quadcopter to track aggressive trajectories, it necessitates the use of an on-board computer. [13] presents an LQR controller based on lie theory. [14] presents an integral action LQR control for a hexacopter UAV, and discusses techniques for tuning the Q and R penalty matrices.

A research gap is present in the literature related to LQR control of UAVs, that can be implemented onboard a microcontroller based flight controller. Most controllers rely on feedback of motor thrust via encoders, or by accurately modelling the motor [15]. This increases accuracy, but also the cost and development time. Multiple studies like [14] [16] [17] [18] rely on low-fidelity MATLAB/SIMULINK simulations for tuning and validating of LQR controller, instead of on a real hardware. This results in significant disparity between simulated results, and between those achieved in real life. Also, the conventional LQR framework may not adequately handle steady-state errors, an issue that is critical in maintaining the precise positioning and orientation of UAVs.

To enhance the performance of traditional LQR, integrating integral action into the control loop presents a promising solution. The augmentation of LQR with integral action (LQRi) aims to effectively mitigate steady-state errors, thereby improving the overall control accuracy of the quadcopter. This integration is especially beneficial in scenarios where external disturbances and model uncertainties are prevalent.

In this paper, we propose an integral action LQR (LQRi) based attitude controller for a quadcopter UAV. We discuss the dynamics of the UAV, design of the controller and a comparison between traditional LQR and LQRi based on high-fidelity simulation, and outdoor flight-tests. Ardupilot is used as a base firmware for this controller, providing robust integration with multiple sensors, while also providing robust state-estimation through an Extended-Kalman-Filter. The code related to LQRi controller implementation has been made

available in this github pr.[1]

## II. MATHEMATICAL MODEL

Understanding the dynamics of a quadcopter is crucial for designing control algorithms, as these dynamics dictate how the quadcopter responds to control inputs. Here, the kinematics and dynamics of the quadcopter is derived based on the quaternion representation of vehicle attitude. This is done because unlike euler angles or rotational matrices, quaternions do not suffer from gimbal lock [19]. Open-source flight controllers like ardupilot rely on angle-axis representation of quaternions for attitude control, similar to [10], which will be used in this controller also. Below is a brief overview of the fundamental dynamics and equations that govern a quadcopter's motion.

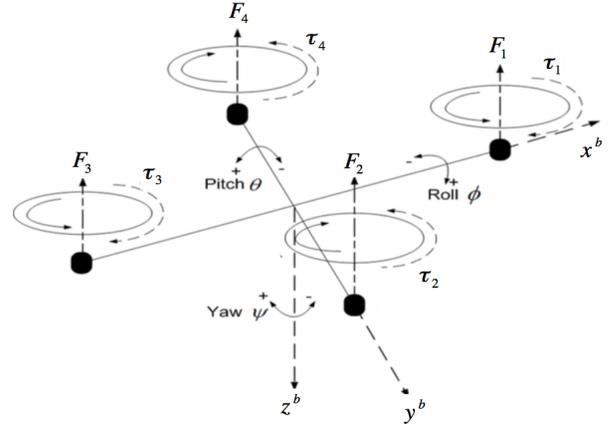

Fig. 1: Representation of quadcopter body frame [22]

### A. Quaternion Mathematics

A quaternion is a four-dimensional number that can represent a body's orientation in three-dimensional space [20] [21]. It is composed of one real part and three imaginary parts and is typically written as:

$$q = q_0 + q_1 i + q_2 j + q_3 k \quad (1)$$

Where $q_0$ corresponds to the real part of the quaternion, $q_1$, $q_2$, and $q_3$ correspond to the vector part, and $i$, $j$, and $k$ are the basis vectors. Quaternions are used for representing 3-dimensional rotation and are heavily used in aerospace as they are not susceptible to singularity, unlike rotation matrices or Euler angles. Norm of a quaternion is analogous to the length of a vector, and is defined as

$$||q|| = \sqrt{q_0^2 + q_1^2 + q_2^2 + q_3^2} \quad (2)$$

The conjugate of a quaternion $q$ is represented as $q^*$ and is defined as:

$$q^* = q_0 - q_1 i - q_2 j - q_3 k \quad (3)$$

An important operation in quaternion algebra is the multiplication of two quaternions $p$ and $q$ with each other. It is defined as the kronecker product of $p$ with $q$:

$$S = p \otimes q = \begin{pmatrix} p_0 q_0 & p_0 q_1 & p_0 q_2 & p_0 q_3 \\ p_1 q_0 & p_1 q_1 & p_1 q_2 & p_1 q_3 \\ p_2 q_0 & p_2 q_1 & p_2 q_2 & p_2 q_3 \\ p_3 q_0 & p_3 q_1 & p_3 q_2 & p_3 q_3 \end{pmatrix} \quad (4)$$

The product of two quaternions represents the combined rotation of the body, first by $p$ and then by $q$. Similarly, the difference between the quaternion $p$ and $q$ is given by:

$$E = p \otimes q^* \quad (5)$$

[1]https://github.com/ArduPilot/ardupilot/pull/26469

Quaternions can be converted to angle-axis representation, which describes a rotation by an angle around a specific axis. For a unit quaternion, this is represented as follows:

$$\theta = 2 \arccos(q_0)$$
$$\vec{Q_a} = \frac{\theta}{\sqrt{1 - q_0^2}} (q_1 i + q_2 j + q_3 k) \quad (6)$$
$$\vec{Q_a} = \vec{\omega_B}$$

where $\theta$ is the rotation angle, $Q_a$ is the vector along the axis of rotation and $\omega_B$ is the angular velocity in the body frame.

### B. Dynamics

Before discussing dynamics, the frame of reference of different vectors needs to be specified
- **Inertial Frame** (Earth-fixed frame): Vectors denoted with subscript $\{I\}$.
- **Body Frame** (Quadcopter-fixed frame): Vectors denoted with subscript $\{B\}$.

The rotational and translational dynamics of the quadcopter can be derived using the Newton-Euler equations of motion for rigid-body, as given in [23], [24] and [19].

$$\ddot{\vec{r}}_I = \vec{a_I}$$
$$\vec{a_I} = q * \frac{F_B}{m} * q^* + \vec{g}$$
$$\dot{q}_I = \frac{1}{2} * \begin{bmatrix} 0 \\ \omega_B \end{bmatrix} \otimes q_I \quad (7)$$
$$\dot{\omega}_B = J^{-1}(\tau_B - \omega_B \times J\omega_B)$$
$$F_B = \sum k_{T_i} * \Omega_i^2$$
$$\tau_B = \begin{bmatrix} k_T * l(\Omega_1^2 - \Omega_2^2 - \Omega_3^2 + \Omega_4^2) \\ k_T * l(\Omega_1^2 - \Omega_2^2 + \Omega_3^2 - \Omega_4^2) \\ k_\tau * l(\Omega_1^2 + \Omega_2^2 - \Omega_3^2 - \Omega_4^2) \end{bmatrix}$$

Here, $r_I$ is the position, $a_I$ is the acceleration, $q$ is the vehicle attitude in the inertial frame and $\Omega_i$ is the rotational speed of propeller i. $F_B$ is the thrust and $\tau_B$ is the torque generated by the motors. $k_T$ and $k_\tau$ are the thrust and torque constants for motor, and $J$ is the moment of inertia of the UAV. Since we are only concerned with control of UAV attitude in this paper, we can reduce the state-space to following states:

$$\mathbf{x} = [Q_a, \omega_B] \quad (8)$$

Where $Q_a$ is the axis angle vector for quaternion attitude. Based on (7) and (6), the reduced dynamics can be given as

$$\begin{bmatrix} \dot{Q}_a \\ \dot{\omega}_B \end{bmatrix} = \begin{bmatrix} \omega_B \\ J^{-1}(\tau_B - \omega_B \times J\omega_B) \end{bmatrix} \quad (9)$$

A state space can be constructed by linearizing the attitude dynamics about hover condition ($\mathbf{x} = 0_{1\times 6}$). The state space can be represented as:

$$\dot{\mathbf{x}} = A\mathbf{x} + B\mathbf{u}$$

$$A = \begin{bmatrix} 0 & 0 & 0 & 1 & 0 & 0 \\ 0 & 0 & 0 & 0 & 1 & 0 \\ 0 & 0 & 0 & 0 & 0 & 1 \\ 0 & 0 & 0 & 0 & 0 & 0 \\ 0 & 0 & 0 & 0 & 0 & 0 \\ 0 & 0 & 0 & 0 & 0 & 0 \end{bmatrix} \quad (10)$$

$$B = \begin{bmatrix} 0 & 0 & 0 \\ 0 & 0 & 0 \\ 0 & 0 & 0 \\ 1/I_{xx} & 0 & 0 \\ 0 & 1/I_{yy} & 0 \\ 0 & 0 & 1/I_{zz} \end{bmatrix}$$

Here, $I_{xx}$, $I_{xx}$, $I_{xx}$ are the moment of inertia of the UAV about $x$, $y$ and $z$ axes respectively, and $\mathbf{u} = \tau_{ref}$ is the required torque for attitude control.

## III. ATTITUDE CONTROL

### A. Linear Quadratic Regulator

Linear Quadratic Regulator (LQR) is a full-state feedback controller that minimizes a quadratic cost function, for a system governed by linear differential equations [25] [26]. The quadratic cost function that is minimized by LQR is given by [27]:

$$J_{cost} = \int_0^\infty (\mathbf{x}^T \mathbf{Q} \mathbf{x} + \mathbf{u}^T \mathbf{R} \mathbf{u}) dt \quad (11)$$

where:
- $\mathbf{x}$ is the vector containing all the states of the system (attitude and body rates).
- $\mathbf{u}$ is the control input vector (desired torques in each axis).
- $\mathbf{Q}$ is a matrix that penalizes errors between reference values and current values of states.
- $\mathbf{R}$ is a matrix that penalizes control effort.

The goal is to determine the control input $\mathbf{u}$ that minimizes this cost function. The solution to the LQR problem involves solving a matrix differential equation known as the Riccati equation. For a state-space as described in (10), the optimal control law $\mathbf{u}$ can be expressed as:

$$\mathbf{u} = -\mathbf{K}(\mathbf{x} - \mathbf{x_{ref}}) \quad (12)$$

where the gain matrix $\mathbf{K}$ is derived from lqr(sys,Q,R) function in MATLAB.

### B. Linear Quadratic Regulator with Integral action

Linear Quadratic Regulator (LQR) with Integral Action, often referred to as Integral LQR (LQR-I), enhances the standard LQR by incorporating integral control. This addition aims to eliminate steady-state errors and improve the system's response to modelling uncertainties or disturbances. In LQRi, the state-space model is augmented to include integral terms. Given a system described as (10), the augmented system can be described as:

$$\dot{\mathbf{x}}_{aug} = \mathbf{A}_{aug}\mathbf{x}_{aug} + \mathbf{B}_{aug}\mathbf{u}$$

$$\mathbf{x}_{aug} = \begin{bmatrix} \mathbf{z} \\ \mathbf{x} \end{bmatrix} \quad (13)$$

where $\mathbf{z}$ is the integral of the attitude errors. The augmented matrices $\mathbf{A}_{aug}$ and $\mathbf{B}_{aug}$ are defined as:

$$\mathbf{A}_{aug} = \begin{bmatrix} \mathbf{A} & \mathbf{0}_{3\times 3} \\ \mathbf{0}_{3\times 6} & \mathbf{I}_{3\times 3} \end{bmatrix}$$

$$\mathbf{B}_{aug} = \begin{bmatrix} \mathbf{0}_{3\times 3} \\ \mathbf{B} \end{bmatrix} \quad (14)$$

## IV. RESULTS AND DISCUSSION

The controller was first validated in Ardupilot's software-in-the loop environment, and then was deployed on a pixhawk cube-orange flight controller for flight testing. The following parameters were used for the design of both LQR and LQRi controllers:

$$Q_{lqr} = diag(0.135_{1\times 3}, 0.0005_{1\times 3})$$
$$Q_{lqri} = diag(0.001, 0.002, 0.001, 0.135_{1\times 3}, 0.0005_{1\times 3})$$
$$R = I_{3\times 3}$$
$$J = \begin{bmatrix} 0.01 & 0 & 0 \\ 0 & 0.02 & 0 \\ 0 & 0 & 0.01 \end{bmatrix}$$
(15)

Here, $J$ is the inertia matrix of the UAV about center of mass. The full-state feedback gain for LQR and LQRi controllers is as follows:

$$K_{lqr} = \begin{bmatrix} 0.387 & 0 & 0 & 0.08 & 0 & 0 \\ 0 & 0.387 & 0 & 0 & 0.08 & 0 \\ 0 & 0 & 0.387 & 0 & 0 & 0.08 \end{bmatrix},$$

$$K_{lqri} = \begin{bmatrix} 0.03 & 0 & 0 & 0.375 & 0 & 0 & 0.09 & 0 & 0 \\ 0. & 0.08 & 0 & 0 & 0.375 & 0 & 0 & 0.125 & 0 \\ 0 & 0 & 0.01 & 0 & 0 & 0.40 & 0 & 0 & 0.09 \end{bmatrix}$$
(16)

### A. SITL simulation

ArduPilot's custom control library greatly facilitated the deployment of LQR control on board the flight controller, as it provided direct interfacing with sensor fusion, motor mixing and flight mode library without disturbing the codebase. It also allowed switching between custom control, and ardupilot's own onboard control, hence allowing safe testing. The C++ code for LQR control was added into CC library. SITL offers more realistic results for following reasons:

- **High Fidelity Simulation:** ArduPilot SITL provides a high-fidelity simulation of the vehicle's dynamics, sensors, and environment. It models the quadcopter's behavior accurately, including factors such as aerodynamics, gravity, and sensor noise.
- **Integration with ArduPilot:** SITL seamlessly integrates with the ArduPilot flight control software. This means that you can use the same codebase for both simulation and real hardware, ensuring consistency and reducing development time.
- **Flexible Configurability:** Users can configure various parameters of the simulated vehicle, such as vehicle type, sensor types, GPS location, and weather conditions. This flexibility allows for a wide range of testing scenarios.
- **Scriptable Interface:** SITL offers a scriptable interface, making it possible to automate testing and create complex simulation scenarios. Users can write scripts to simulate specific flight situations, sensor failures, or environmental changes.
- **Support for Multiple Vehicle Types:** In addition to quadcopters, SITL supports various vehicle types, including fixed-wing aircraft, rovers, and submarines. This versatility makes it suitable for a wide range of robotic applications.
- **Realistic Sensor Models:** SITL includes realistic sensor models for GPS, IMU (Inertial Measurement Unit), barometer, magnetometer, and more. These models mimic the behavior of actual sensors, allowing for sensor fusion and navigation testing.

A series of mission were simulated in SITL for testing the LQR controller, in which waypoint commands were given for the UAV to track. The results of the simulation can be summarized below (all results are in degrees):

| Axis | MSE | RMSE | Mean Deviation |
|---|---|---|---|
| Roll | 1.056 | 1.027 | 0.28 |
| Pitch | 0.98 | 0.99 | 0.27 |
| Yaw | 8.75 | 2.95 | 0.91 |

TABLE I: LQR SITL tracking performance errors

| Axis | MSE | RMSE | Mean Deviation |
|---|---|---|---|
| Roll | 0.68 | 0.83 | 0.18 |
| Pitch | 0.42 | 0.65 | 0.16 |
| Yaw | 4.87 | 2.2 | 0.65 |

TABLE II: LQRi SITL tracking performance errors

### B. Flight testing

The quadcopter for flight testing was based on dji-F450 frame. 920kv brushless motors were used for propulsion, along with 10 inch propellers. Pixhawk cube-orange+ was used as the flight controller, and the system was powered by a 3s 4200mah LiPo battery.

The tracking accuracy for LQR and LQRi is as follows:

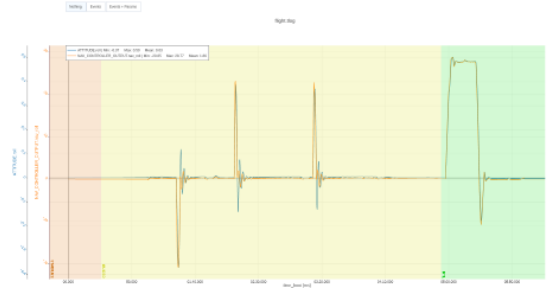

(a) Roll response of LQR

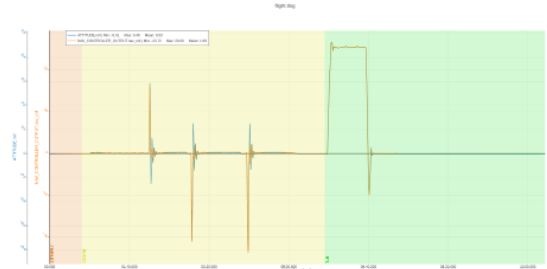

(b) Roll response of LQRi

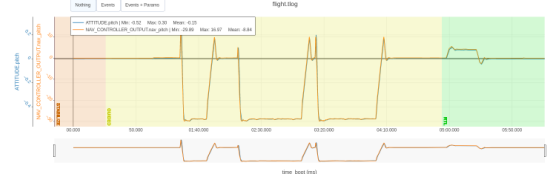

(c) pitch response of LQR

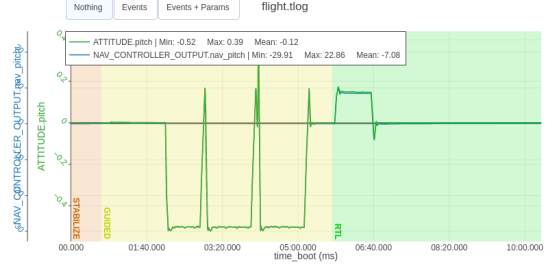

(d) pitch response of LQRi

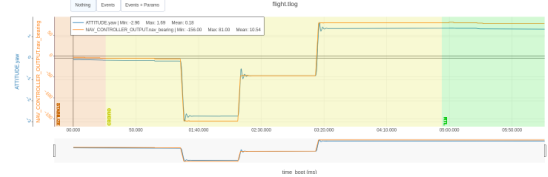

(e) yaw response of LQR

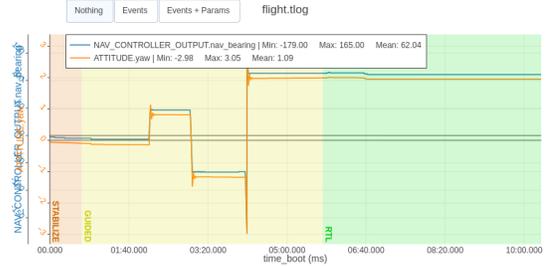

(f) yaw response of LQRi

Fig. 2: Response of attitude controllers in SITL

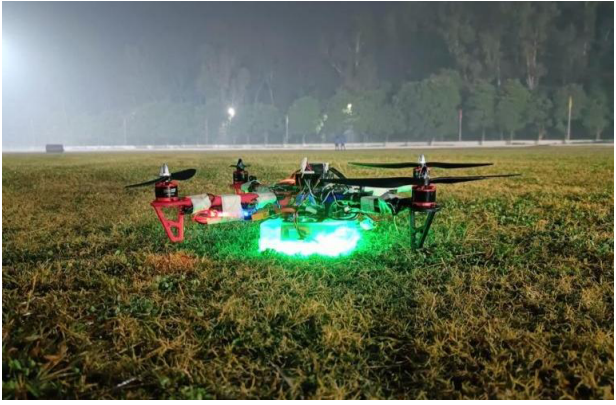

Fig. 3: Quadcopter testbed

| Axis  | MSE   | RMSE  | Mean Deviation |
|-------|-------|-------|----------------|
| Roll  | 16.69 | 4.085 | 3.09           |
| Pitch | 7.4   | 2.68  | 1.28           |
| Yaw   | 31.28 | 5.59  | 3.07           |

TABLE III: LQR attitude tracking performance

| Axis  | MSE   | RMSE  | Mean Deviation |
|-------|-------|-------|----------------|
| Roll  | 3.45  | 1.85  | 1.29           |
| Pitch | 2.35  | 1.53  | 1.00           |
| Yaw   | 23.39 | 4.83  | 3.61           |

TABLE IV: LQRi attitude tracking performance

The parameters shown in table III and table IV clearly show that LQR with integral action is better at tracking reference attitude commands, with a much better performance in roll, pitch and yaw axis

## V. Conclusion and Future Work

This paper proposed an integral action LQR attitude control for a Quadcopter UAV, developed from a quaternion-based mathematical model. The LQR gain for full state feedback was calculated and the controller was tested and validated for stable flight in a software-in-the-loop environment. The LQRi controller showed an improvement of over 30% against traditional LQR in attitude reference tracking. Flight tests were conducted onboard a dji-f450 quadcopter platform with Pixhawk Cube Orange+ as a flight controller. LQRi performed 40% better in trajectory tracking than traditional LQR in roll, and pitch axis, while being 13% better in the yaw axis. However, in the case of the yaw axis, the results showed a sensitivity of the controller to Q and R matrices. Hence, an autonomous tuning method will be investigated for tuning these hyperparameters in future studies. Also, a method to track body rate commands instead of just regulating them, free of information about motor parameters will be investigated in future.

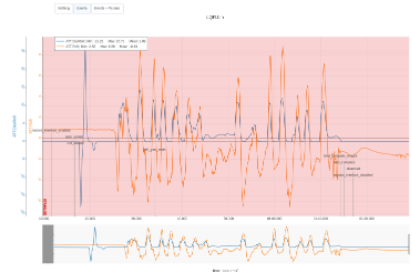

(a) Roll response of LQR

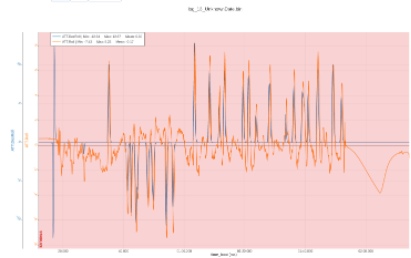

(b) Roll response of LQRi

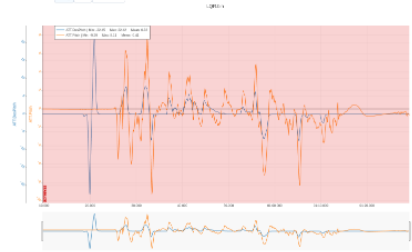

(c) pitch response of LQR

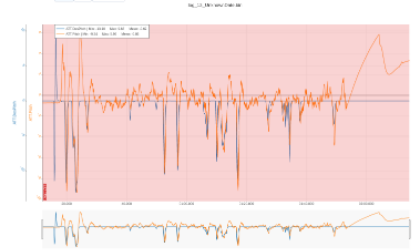

(d) pitch response of LQRi

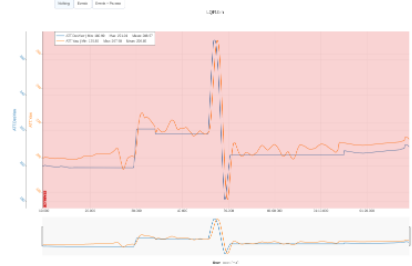

(e) yaw response of LQR

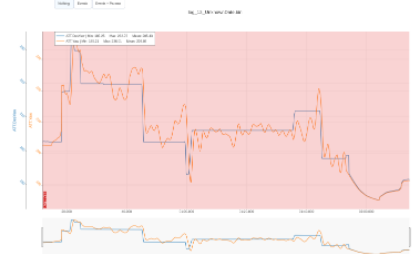

(f) yaw response of LQRi

Fig. 4: Flight test response of attitude controllers